# Developing a New Approach for Arabic Morphological Analysis and Generation

Mourad Gridach (1), Noureddine Chenfour (2)
(1) *Mathematics and Computer Science Department, Sidi Mohamed Ben Abdellah University – Faculty of Sciences, Fez, Morocco. E – Mail: mourad_i4@yahoo.fr*
(2) *Mathematics and Computer Science Department, Sidi Mohamed Ben Abdellah University – Faculty of Sciences, Fez, Morocco. E – Mail: chenfour@yahoo.fr*

## Abstract

Arabic morphological analysis is one of the essential stages in Arabic Natural Language Processing. In this paper we present an approach for Arabic morphological analysis. This approach is based on Arabic morphological automaton (AMAUT). The proposed technique uses a morphological database realized using XMODEL language. Arabic morphology represents a special type of morphological systems because it is based on the concept of scheme to represent Arabic words. We use this concept to develop the Arabic morphological automata. The proposed approach has development standardization aspect. It can be exploited by NLP applications such as syntactic and semantic analysis, information retrieval, machine translation and orthographical correction. The proposed approach is compared with Xerox Arabic Analyzer and Smrz Arabic Analyzer.

## 1. Introduction

Nowadays, Arabic language faces many challenges. The first important challenge is the requirement to analyze Arabic morphology with high quality because it is considered as the essential stage in many NLP applications such as Information Retrieval and Machine Translation. The second challenge is concerning the use of morphology in machine translation systems. Koehn and Hoang (2007) have shown that factored translation models containing morphological information lead to better translation performance. Morphological analysis becomes more important when translating to or from morphologically rich languages such as Arabic. The third challenge is that morphological analysis is considered as the first step before syntactic analysis.

Arabic is a morphologically complex language. The morphological analysis of any word consists of determining the values of a large number of features, such as basic part-of-speech (i.e., noun, verb, etc.), gender, person, number, voice, information about the clitics, etc. (Habash, 2005). There has been much work on Arabic morphology (see Al-Sughaiyer and Al-Kharashi, 2004). Since, lots of morphological analysis approaches are available now, some of them have a commercial purpose and the others are available for research and evaluation (Attia, 2006).

In this paper we present an approach for Arabic morphological analysis based on Arabic morphological automaton technique. To construct an Arabic morphological automaton, we used particularities of Arabic morphology that are concretized on multilevel: verbs and nouns are also characterized by a specific representation named the matrix "root – scheme". Arabic nouns and verbs are derived from roots by applying schemes to these roots to generate Arabic stems and then adding prefixes and suffixes to the stems to form a correct word in Arabic language. Table 1 show some schemes applied to the root "ktb" (كتب). Our approach is used to develop a morphological analyzer for Arabic language.

Table 1. Some examples of schemes to generate stems from the root "ktb" (كتب)

| **Scheme** | Facal | fAcil | mafcUl | Mafcal | ficAl |
|---|---|---|---|---|---|
| **Stem generated** | Katab | kAtib | MaktUb | Maktab | kitAb |

The structure of the article is as follows. First, in the introduction we discuss the challenges of Arabic language and the importance of morphological analysis as an essential step in Natural Language Processing. We present an overview of Arabic morphology in the second section. We discuss some Arabic morphological analysis approaches related to the presented work in the third section. In the fourth section, we present our lexicon. In section five, we present our approach for Arabic morphological analysis. We present the application of the presented work in Arabic morphological analysis in section six. In the seventh section, we evaluate the proposed technique. In section eight, we discuss the obtained results. Finally, in the last section, we draw some conclusions.

## 2. Overview of Arabic morphology

Morphology is the branch of linguistics that deals with the internal structure of words. It studies word formation, including affixation behavior, roots, and pattern properties. Morphology can be classified as either inflectional or derivational. Inflectional morphology is applied to a given stem with predictable formation. It does not affect the word's grammatical category, such as noun, verb, etc. Case, gender, number, tense, person, mood, and voice are some examples of characteristics that might be affected by inflection. Derivational morphology, on the other hand, concatenates to a given word a set of morphemes that may affect the syntactic category of the word. The distinction between these two classes is not an easy one to make, and it differs from one language to another. In this section, we deal with Arabic morphology.

The origin of Arabic is very different from the other languages especially European languages. It includes 28 letters and they are all considered as consonants. The Arabic writing system is very different from the most of the other languages because it is written from right to left. Arabic morphological representation is rather complex because of some morphological phenomenon like the agglutination phenomenon. Letters change forms according to their position in the word (beginning, middle, end and separate). Table 2 gives an example of different forms of the letter "g" /الغين/ at different positions.

Table 2. The letter "g" /**الغين**/ representation in the different position

| Beginning | Middle | End | Separate |
|---|---|---|---|
| غـ | ـغـ | ـغ | غ |

It seems that Arabic traditional grammarians (Ibrahim, 2002) have been persuaded by morphology to classify words into only three types: verbs, nouns and prepositions and particles. Adjectives take almost all the morphological forms of nouns. Compared to other languages, Arabic is mainly derivational while others are concatenative. It is characterized by its very rich derivational morphology, where almost all of words are derived from roots by applying patterns (Darwish, 2002). There is in Arabic around 10000 roots, which are in general made up of three, four, or five letters. The roots with three letters alone generate approximately 85% of the Arabic words (De Roeck, 2000). Concerning the Arabic morphemes, they fall into three categories: templatic morphemes, affixes morphemes, and non-templatic word stems (NTWSs). NTWSs are word stems that are not constructed from a "root – scheme" combination. Verbs are never NTWSs (Habash, 2006).

In this presented work, we have classified the first category of morphemes in two types which are needed to create an Arabic word stem: roots and schemes. The root morpheme is an ordered sequence of valid three or four characters from the alphabet and rarely five characters. The root is not a valid Arabic word (for example /ktb/ /كتب/). The scheme morpheme is an ordered sequence of characters. Some of these characters are constants and some are variables. The variables characters are to be substituted with the characters of an Arabic root to generate a word called the "stem". There are different schemes for the triliteral and tetraliteral roots. Table 3 shows schemes for some Arabic verbs classified as original schemes. Note that the scheme is not a valid Arabic word, whereas the stem is a valid word. The word stem is formed by substituting characters of the root into certain verb schemes. The following example shows an example of constructing an Arabic word stem using the root and scheme.

Root "ktb" /ك ت ب/
Scheme "tafAcala" /تَفَاعَلَ/
Stem "takAtaba" /تَكَاتَبَ/

Table 3. Some schemes for some Arabic verbs

| Facala | eiftacala | einfacala | faclala | eafcala | fAcala | eistafcala | tafAcala |
|---|---|---|---|---|---|---|---|
| فَعَلَ | افْتَعَلَ | انْفَعَلَ | فَعْلَلَ | أَفْعَلَ | فَاعَلَ | اسْتَفْعَلَ | تَفَاعَلَ |

For the affixes morphemes, they can be added before or after a root or a stem as a prefix or suffix. For prefixes, they can be added before the stem like the prefix "sa" /س/ which used to express the future while suffixes can be added after the stem like the suffix "Una" /ون/. In this paper, we use this classification of Arabic morphemes to concretize our approach for Arabic morphological analysis.

# 3. Related work

Much work has been done in the area of Arabic morphological analysis and generation in a variety of approaches and at different degrees of linguistics depths (Al-Sughaiyer and Al-Kharashi, 2004). The most of the approaches tend to target a specific application (Khoja, 2001; Darwish, 2002; Diab et al., 2007a). The most referenced systems are works done by Habash et al., Smrz, Buckwalter and Beesley. They are available for research and evaluation and well documented. In this section, we discuss these works to clarify their working method and approach in Arabic morphological analysis.

### 3.1 MAGEAD: A Morphological Analyzer and Generator for Arabic Dialects

MAGEAD is one of the existing morphological analyzers for the Arabic language available for research. It's a functional morphology systems compared to Buckwalter morphological analyzer which models form-based morphology (M. Altantawy et al., 2010). To develop MAGEAD, they use a morphemic representation for all morphemes and explicitly define morphophonemic and orthographic rules to derive the allomorphs. The lexicon is developed by extending Elixir-FM's lexicon. The advantage of this analyzer is that it processes words from the morphology of the dialects which they considered as a novel work in this domain, but unfortunately this analyzer needs a complete lexicon for the dialects to make the evaluation more interesting and convincing, and to verify these claims.

### 3.2 ElixirFM: an Arabic Morphological Analyzer by Otakar Smrz

ElixirFM is an online Arabic Morphological Analyzer for Modern Written Arabic developed by Otakar Smrz available for evaluation and well documented. This morphological analyzer is written in Haskell, while the interfaces in Perl. ElixirFM is inspired by the methodology of Functional Morphology (Forsberg and Ranta, 2004) and initially relied on the re-processed Buckwalter lexicon (Buckwalter, 2002). It contains two main components: a multi- purpose programming library and a linguistically morphological lexicon (Smrz, 2007). The advantage of this analyzer is that it gives to the user four different modes of operation (Resolve, Inflect, Derive and Lookup) for analyzing an Arabic word or text. But the system is limited coverage because it analyzes only words in the Modern Written Arabic.

### 3.3 Buckwalter Arabic Morphological Analyzer

This analyzer is considered as one of the most referenced in the literature, well documented and available for evaluation. It is also used by Linguistic Data Consortium (LDC) for POS tagging of Arabic texts, Penn Arabic Treebank, and the Prague Arabic Dependency Treebank (Atwell et al., 2004). It takes the stem as the base form and root information is provided. This analyzer contains over 77800 stem entries which represent 45000 lexical items. However, the number of lexical items and stems makes the lexicon voluminous and as result the process of analyzing an Arabic text becomes long.

### 3.4 Xerox Arabic Morphological Analysis and Generation

Xerox Arabic morphological Analyzer is well known in the literature and available for evaluation and well documented. This analyzer is constructed using Finite State Technology (FST) (Beesley, 1996 and 2000). It adopts the root and pattern approach. Besides this, it includes 4930 roots and 400 patterns, effectively generating 90000 stems. The advantages of this analyzer are, on the one hand, the ability of a large coverage. On the other hand, it is based on rules and also provides an English glossary for each word. But the system fails because of some problems such as the overgeneration in word derivation, production of words that do not exist in the traditional Arabic dictionaries (Darwish, 2002) and we can consider the volume of the lexicon as another disadvantage of this analyzer which could affect the analysis process.

# 4. Lexicon

The lexicon is the set of valid lexical forms of a language. As in any morphological analysis approach, the enhancement level of the lexicon determines the quality of the analysis. There are two aspects that contribute to this enhancement level. The first aspect concerns the number of lexicon entries contained in the lexicon. Second aspect concerns the richness in linguistics information contained by the lexicon entries. We mention that large Arabic morphological analyzers used the BAMA lexicon. It was used in the creation of Elixir-FM. Also MAGEAD used it by extending Elixir-FM's lexicon.

There has been much work to construct an Arabic lexicon with the optimal way. For an overview see (Al-Sughaiyer and Al-Kharashi, 2004). Currently, another method is used for representing, designing and implementing the lexical resource. This method used the Lexical Markup Framework (LMF). It was used in lots of languages (Indo-European), but for Arabic language this method still in progress towards a standard for representing the Arabic linguistic resource. Lexical Markup Framework (LMF, ISO-24613) is the ISO standard which provides a common standardized framework for the construction of natural language processing lexicons. The US delegation is the first which started the work on LMF in 2003. In early 2004, the ISO/TC37 committee decided to form a common ISO project with Nicoletta Calzolari (Italy) as convenor and Gil Francopoulo (France) and Monte George (US) as editors (Francopoulo and George, 2008).

To represent Arabic morphological knowledge with the optimal way, we conceived an innovative language adapted for this specific situation: it is the XMODEL language (XML-based MOrphological DEfinition Language). As a result, all morphological entries are gathered in an XMODEL files. Using the new language helps direct search for information. It allows representing the whole components, properties and morphological rules with a very optimal way. To clarify the last point, we note that our morphological database contains 960 lexicon items (morphological components) and 455 morphological rules to be applied to these morphological components which present a remarkable reduction in the number of entries in the lexicon compared to the existing systems (Xerox and Buckwalter). This representation helps us achieve the following goals:

- ✓ A symbolic definition, declarative and therefore progressive of the Arabic morphology.
- ✓ A morphological database independent of processing that will be applied (see later).
- ✓ A considerable reduction of the number of morphological entries.
- ✓ The scheme allows defining the maximum morphological components by means of XMODEL language.

Our language makes it possible to represent Arabic morphology as morphological classes and rules. Accordingly, our Arabic morphological database will be composed of three parts: morphological classes, morphological properties and morphological rules.

The next paragraph presents XMODEL language which allows representing Arabic morphological knowledge and consists of three parts.

### 4.1 Morphological component class

It allows representing all Arabic morphological components. It also permits to gather a set of morphological components having the same nature, the same morphological characteristics and the same semantic actions. Relying on the notion of scheme "*ealwazn*" /الوزن/, this class allows a better optimization hence, a considerable reduction of morphological entries. Figure 1 shows a morphological component class representing four schemes ("*facala*", "*facila*", *facula*" and *faclala*"). These schemes are called original schemes. In this morphological component class, they are considered as morphological components.

```
  <?xml version="1.0" encoding="ISO-8859-1" ?>
- <package name="OrigineSchemesPackage">
  - <morphological_class name="OriginSchemeS
    - <properties>
        <modifier>final</modifier>
        <is>FinalVerbS</is>
        <is>Number.NSg</is>
        <is>Person.Pr3</is>
        <is>Gender.GMa</is>
      </properties>
      <component name="facala" id="1" />
      <component name="facila" id="2" />
      <component name="facula" id="3" />
      <component name="faclala" id="4" />
    </morphological_class>
  </package>
```

```
  <?xml version="1.0" encoding="UTF-8" ?>
- <package name="OrigineSchemesPackage">
  - <morphological_class name="OriginSchemeS">
    - <properties>
        <modifier>final</modifier>
        <is>Number.NSg</is>
        <is>Person.Pr3</is>
        <is>Gender.GMa</is>
      </properties>
      <component name="فَعَلَ" id="1" />
      <component name="فَعِلَ" id="2" />
      <component name="فَعُلَ" id="3" />
      <component name="فَعْلَلَ" id="4" />
    </morphological_class>
  </package>
```

Figure 1. Representation of some verbs schemes using XMODEL language

### 4.2 Morphological properties class

It allows characterizing the different morphological components represented by the morphological class: a morphological property class contains a set of morphological descriptors or morphological values of properties that would be assigned to different morphological components. We mention, for example, the property "*Gender*" which will distinguish between masculine and feminine components. The morphological properties are not related to a specific morphological class which makes it necessary to define them outside the morphological classes. Figure 2 shows an example of morphological properties class that contains two morphological properties (Person and Gender). Each morphological property contains a set of descriptors.

```
  <?xml version="1.0" encoding="ISO-8859-1" ?>
- <package name="PropertyPackage">
  - <morphological_properties>
    - <property name="Person" type="exclusive">
        <descriptor name="Pr1" />
        <descriptor name="Pr2" />
        <descriptor name="Pr3" />
      </property>
    - <property name="Gender" type="additive">
        <descriptor name="GFe" />
        <descriptor name="GMa" />
      </property>
    </morphological_properties>
  </package>
```

Figure 2. Representation of morphological properties "*Gender*" and "*Person*"

We have added the attribute "*type*" to work out the problem of the semantic of the morphological descriptors that might be **exclusive** (the morphological component can not be characterized by the morphological descriptors of the same property as in the case of the "*Person*" property) or **additive** (the morphological component can be characterized by the morphological descriptors of the same property as it is the case in the "*Gender*" property).

There are two strategies to characterize the morphological components using the properties:

#### 4.2.1 Components property

A morphological class can use a list of morphological descriptors to define its components. Generally speaking; each morphological component can have its own morphological descriptors. As for the "*Gender*" property, some components of this class can be masculine while the others can be feminine. This type of properties is named the **components property**. In order to put them into practice, we have introduced the "*uses*" tag. This means that the different morphological descriptors defined by components property can be used by the different morphological components of the morphological class. Figure 3 shows an example of three components property (Gender, Number and Place). They are used to characterize the two morphological components "*hAvA*" /هذا/ and "*vAlika*" /ذلك/.

```
  <?xml version="1.0" encoding="ISO-8859-1" ?>
- <morphological_class name="NPEichArat">
  - <properties>
      <uses>Gender</uses>
      <uses>Number</uses>
      <uses>Place</uses>
    </properties>
  - <component name="hAvA">
      <md key="NSg" />
      <md key="GMa" />
      <md key="pro" />
    </component>
  - <component name="vAlika">
      <md key="NSg" />
      <md key="GMa" />
      <md key="LOI" />
    </component>
  </morphological_class>
```

```
  <?xml version="1.0" encoding="UTF-8" ?>
- <morphological_class name="NPEichArat">
  - <properties>
      <uses>Gender</uses>
      <uses>Number</uses>
      <uses>Place</uses>
    </properties>
  - <component name="هذا">
      <md key="NSg" />
      <md key="GMa" />
      <md key="pro" />
    </component>
  - <component name="ذلك">
      <md key="NSg" />
      <md key="GMa" />
      <md key="LOI" />
    </component>
  </morphological_class>
```

Figure 3. Components property (« *Gender* » « *Number* » and « *Place* ») characterizing the components "*hAvA*" and "*vAlika*".

### 4.2.2 Classes property

This one requires assigning a set of morphological components the commons morphological properties. For example, all components are masculine names. This type of property is known as **classes property**. To concretize the use of classes property, we introduce the "*is*" tag. Figure 4 shows an example of two classes property ("Number.NSg" and "Gender.GMa"). It means that all the five schemes are singular and have masculine gender. We mention that the same morphological components class can use both of the tags "*uses*" and "*is*".

```xml
<?xml version="1.0" encoding="ISO-8859-1" ?>
- <morphological_class name="OriginSchemeS">
  - <properties>
      <is>Number.NSg</is>
      <is>Gender.GMa</is>
    </properties>
    <component name="facala" id="1" />
    <component name="facila" id="2" />
    <component name="facula" id="3" />
    <component name="faclala" id="4" />
    <component name="eafcala" id="5" />
  </morphological_class>
```

```xml
<?xml version="1.0" encoding="UTF-8" ?>
- <morphological_class name="OriginSchemeS">
  - <properties>
      <modifier>final</modifier>
      <is>Number.NSg</is>
      <is>Gender.GMa</is>
    </properties>
    <component name="فَعَلَ" id="1" />
    <component name="فَعِلَ" id="2" />
    <component name="فَعُلَ" id="3" />
    <component name="فَعْلَلَ" id="4" />
    <component name="أَفْعَلَ" id="5" />
  </morphological_class>
```

Figure 4. Example of classes property

### 4.2.3 Reference property

Another strong point of the XMODEL language is the introducing of the notion of **reference property** which has an important role to benefit from the specificities of the Arabic morphology. As for the Arabic language some morphological components might be conjugated forms of other components which we call original components. An example of these forms is the following components "*afcalu*", "*afcilu* ", "*afculu*" (see figure 6). These components are all conjugated forms of the component "*facala*" (see figure 5). We have specified the reference between components using the "*ref*" tag.

```xml
<?xml version="1.0" encoding="ISO-8859-1" ?>
- <morphological_class name="OriginSchemeS">
    ...
    <component name="facala" id="1" />
    <component name="facila" id="2" />
    <component name="facula" id="3" />
    <component name="faclala" id="4" />
    <component name="eafcala" id="5" />
    ...
  </morphological_class>
```

Figure 5. Example of some verbs schemes

```xml
<?xml version="1.0" encoding="ISO-8859-1" ?>
- <morphological_class name="VerbSainMuDAric">
  - <properties>
      <ref>OriginSchemeS</ref>
    </properties>
    <component name="afcal" key="1" />
    <component name="afcil" key="2" />
    <component name="afcul" key="3" />
    <component name="ufcil" key="5" />
    ...
  </morphological_class>
```

Figure 6. The conjugated forms of some verbs

In order to concretize this reference between components, we have opted the attribute "*id*" to the original component (see figure 5). This attribute is specified in the "*component*" tag. The components that are conjugated forms will use this code as an attribute of that tag (the "*key*" attribute) to indicate this reference (see figure 6).

### 4.3 Morphological rules class

Firstly, it should be noted that we developed 455 morphological rules for the Arabic language. They help us combine some morphological components (morphemes) together to generate correct language words. They use the different morphological components classes as well as the morphological properties classes. The morphological rules classes allow us to add new morphological descriptors which do not belong to the union of morphological descriptors of components of rules. As a result, they are considered as a generator of language words. The implementation of the morphological rules class permits to put into practice all the possible concatenations between components. Figure 7 shows a morphological rules class named "*prefixesSuffixes*" that contains two rules. The two rules allow generating words which begin by the prefix "*la*" and "*bi*".

```
<?xml version="1.0" encoding="ISO-8859-1" ?>
- <package name="RulesPackage">
  - <rules_class name="prefixeSuffixes">
    - <rule>
        <morpheme key="PrefixeHJar.JarMaDmUr" component="la" />
        <morpheme key="DamirMuttaSil.RDamirMuttaSil" />
      </rule>
    - <rule>
        <morpheme key="PrefixeHJar.JarMaDmUr" component="bi" />
        <morpheme key="DamirMuttaSil.JDamirMuttaSil" />
      </rule>
    </rules_class>
  </package>
```

Figure 7. Morphological rules class representing the components prefixed by the prefix “*la*” and “*bi*”

The structuring of our morphological database using XMODEL language allows us to generate the Arabic morphological automaton.

## 5. Our approach

There has been much work on Arabic morphological analysis where lots of approaches are implemented to satisfy that area of research. For an overview of the approaches of Arabic morphological analysis, see (Al-Sughaiyer and Al-Kharashi, 2004).

The method presented is based on Arabic Morphological Automaton (AMAUT). It is considered among the most efficient methods. AMAUT is responsible for both analysis and generation tasks. A word is accepted by an AMAUT if it belongs to a correct word in Arabic. Generally speaking, an Arabic morphological automaton is represented as $<Q, \sum, q_0, F, \tau>$. Where:

- Q is a finite set of states of the control unit which represents the states of an AMAUT.
- $\sum$ is a finite input tape alphabet symbols. For an AMAUT, it is constituted of the Arabic characters.
- $q_0$ is the start state of the AMAUT. It is constituted of only one start state in the case of a morphological automaton.
- F is a subset of Q. It represents the accepting states of the AMAUT. It gives the morphological descriptors (features) that characterize each word analyzed.
- The set $\tau$ also represents the transition function of the AMAUT.

Consequently, implementing AMAUT needs to use the lexicon discussed in the previous section. We have to extract all the morphological rules from the lexicon and implement an AMAUT for each rule. So to realize that implementing, we have to use some operations such as concatenation and union. In the next paragraphs, we explain how we can use these two operations to generate an AMAUT for a definite morphological rule. The following morphological rule (“*rule_1*”) is responsible to product Arabic numbers that accept the suffix “un”.

```
- <package name="RulesPackage">
  - <rules_class name="cardNbCRules">
    - <rule id="rule_1">
        <morpheme key="CardNumber.CNAccepteSCID"/>
        <morpheme key="CasSuffixe.SCID" component="un"/>
        <idp name="CNIndefMarfUc"/>
      </rule>
      ...
    </rules_class>
  </package>
```

So to generate the AMAUT representing this morphological rule (“*rule_1*”), we concatenate the first morpheme (key = “*CardNumber.CNAccepteSCID*”), which represents Arabic numbers that accept suffixes (like “wAHid” /واحد/, “ca^arat” /عشرة/, “~amAn” /ثمان/, etc.), with the second one (key = “*CasSuffixe.SCD*” component = “*un*”), which represents the suffix “un”. Figure 8 shows the resulting AMAUT obtained from the rule “*rule_1*”.

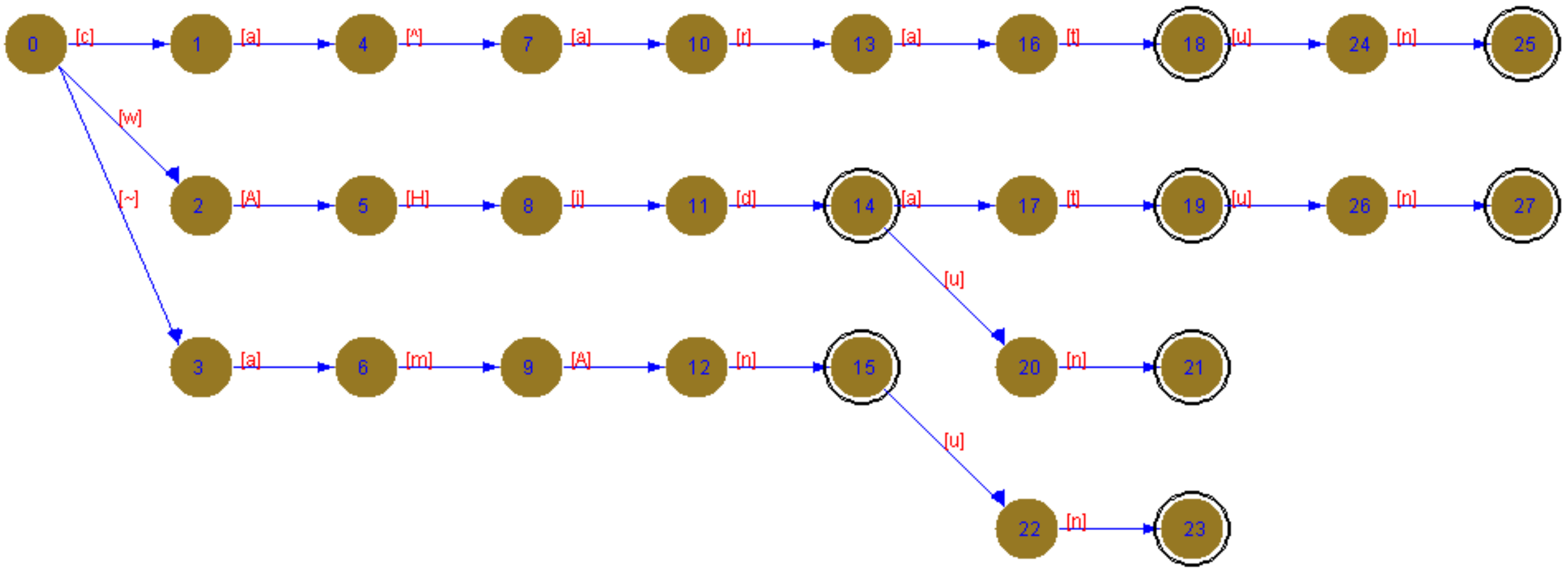

Figure 8. A morphological automaton representing the above morphological rule

In addition to the concatenation used to concatenate morphemes or morphological automata together, we used the union operation to associate two or several morphological automata generated by the first operation, each one represent a definite morphological rule. Figure 9 concretizes the use of the union operation. It contains a morphological rules class (“*cardNbCRules*”) that shows an example of two morphological rules (“*rule_1*” and “*rule_2*”) each one represents the Arabic numbers that accept suffixes which will be concatenated with the two suffixes “un” and “an”.

```
- <package name="RulesPackage">
  - <rules_class name="cardNbCRules">
    - <rule id="rule_1">
        <morpheme key="CardNumber.CNAccepteSCID"/>
        <morpheme key="CasSuffixe.SCID" component="un"/>
        <idp name="CNIndefMarfUc"/>
      </rule>
    - <rule id="rule_2">
        <morpheme key="CardNumber.CNAccepteSCID"/>
        <morpheme key="CasSuffixe.SCID" component="an"/>
        <idp name="CNIndefManSUb"/>
      </rule>
      ...
    </rules_class>
  </package>
```

Figure 9. Morphological rule representing the Arabic numbers which accept suffixes “*un*” and “*an*”

The AMAUT generated by the morphological rule “rule_1” is showed in figure 8. Figure 10 shows an AMAUT generated by the morphological rule “rule_2” which represents the Arabic numbers that accept the suffix “an”.

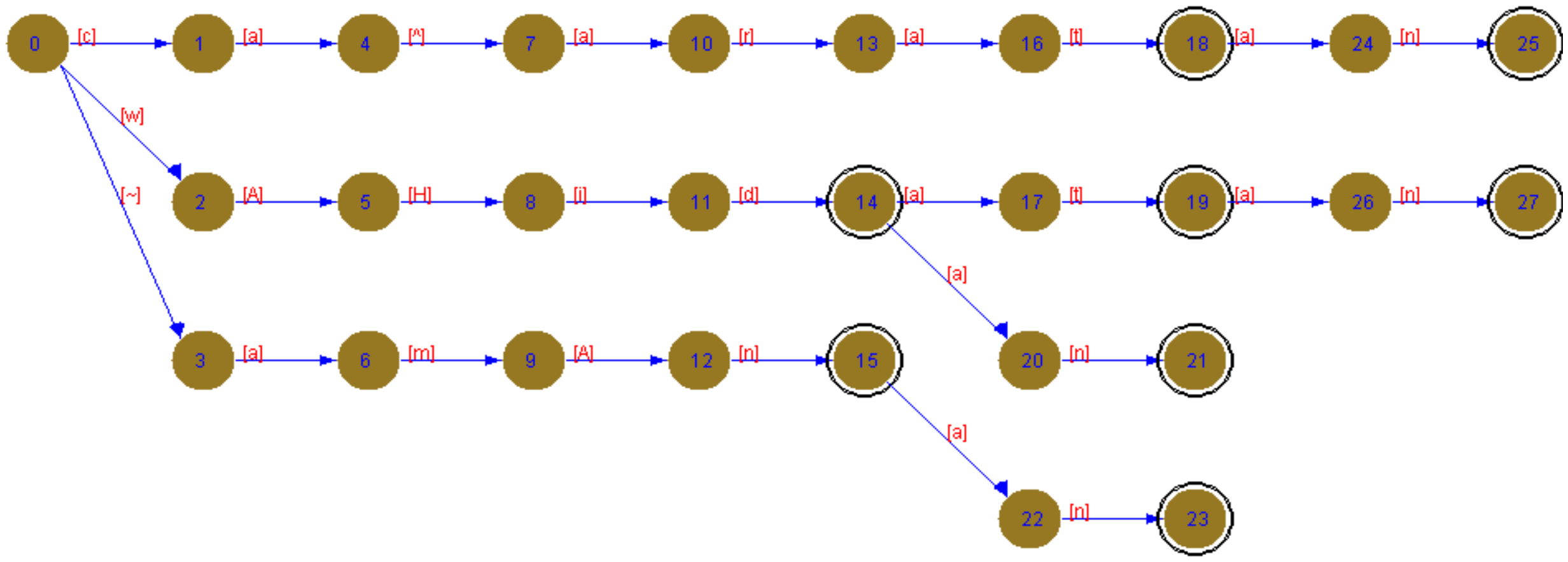

Figure 10. The AMAUT generated by the morphological rule “rule_2”

In figure 9, we have two morphological rules; each one generates an AMAUT. We used the union operation to associate the first AMAUT (represented in figure 8) which represents the rule identified by “*rule_1*” with the second AMAUT (represented in figure 10) which represents the rule identified by “*rule_2*”. Figure 11 shows the resulting AMAUT.

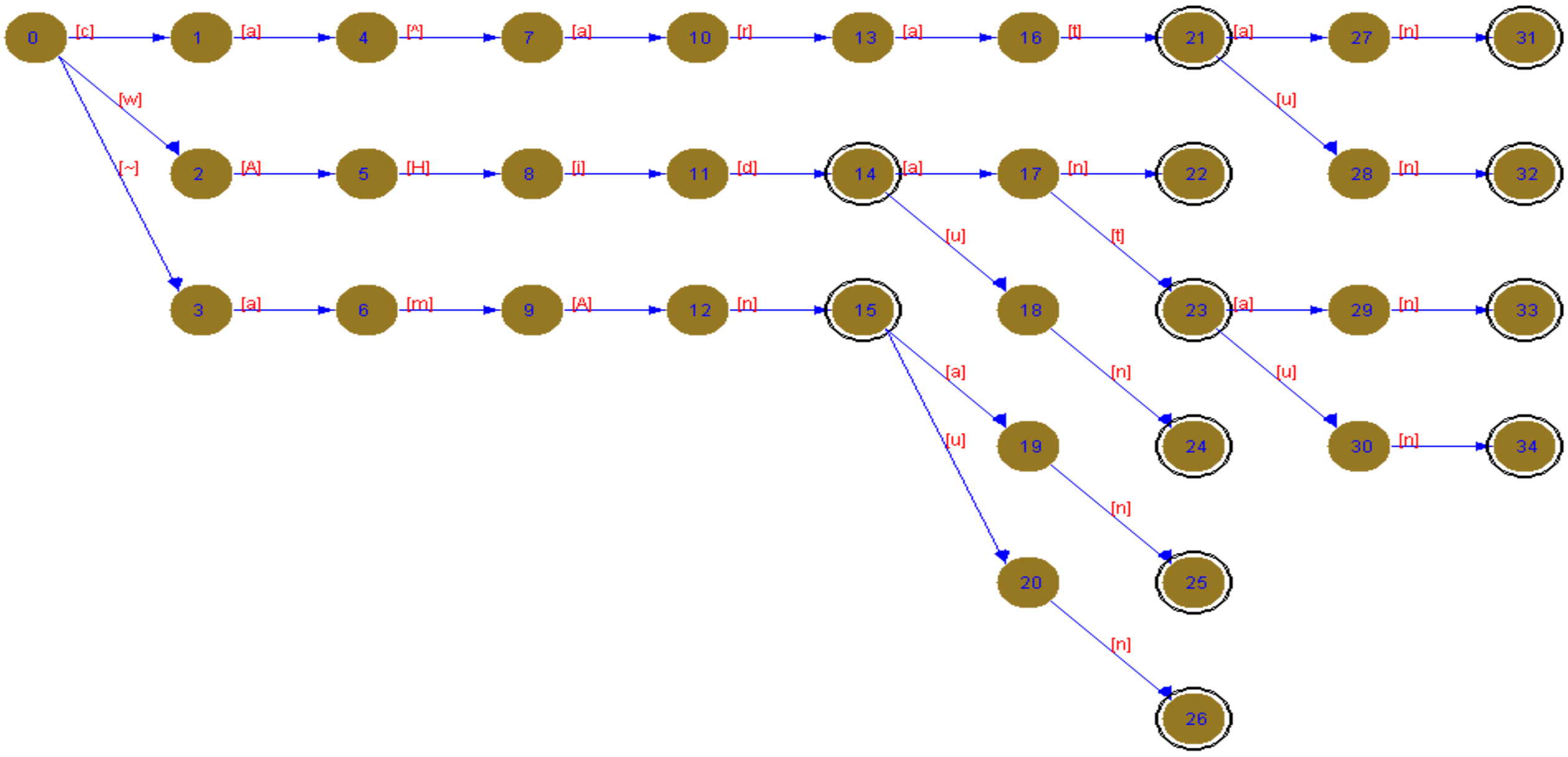

Figure 11. The AMAUT generated by the morphological rule in figure 9

In the following paragraphs, we present a detail of how to implement all the AMAUT and the technique used in the implementation.

So as to implement an AMAUT, we have classified Arabic words in to two categories: the first category is that which submits to the derivation process, while the second one doesn’t. This derivation process is generated by a set of morphological rules known in the Arabic grammar under the name “*qawAcidu eaSSarfi*” /قواعد الصرف/. They repose on the manipulation of a set of very determined schemes named “*ealeawzAn*” /الأوزان/.

The scheme (measure or form) is a general mould composed of an ordered sequence of characters. Some of these characters are constants (instantiated) and some are variables (uninstantiated) (El-Sadany, 1989). The uninstantiated characters are to be substituted (instantiated) with the characters of an Arabic root to generate a word called the “stem.” There are different schemes for the trilateral and tetraliteral roots. Note that the scheme is not a valid Arabic word, whereas the stem is a valid word.

At the graphical level, a scheme generally constitutes of (Tahir et al., 2004):

- Three consonants that are represented by the letters “f” /ف/, “c” /ع/ and “l” /ل/ with a possibility to duplicate the last letter “l” as in the schemes case that correspond to a four letters root like “*faclala*” /فعلل/.
- Some consonants that serve as tools to extend the root like “*stafcala*” /استفعل/ and “*tafAcala*” /تفاعل/.
- An adequate vowels group.

We have grouped in the first category of words the following items:

- Derived nouns “*ealaSmAe ealmu^taqqa*” /الأسماء المشتقة/.
- Strong verbs “*ealeafcAl eaSSaHiyHa*” /الأفعال الصحيحة/: these are the verbs that contain no weak letters. In the Arabic language, there are three weak letters: “w” /و/, “A” /ا/ and “y” /ي/.
- Weak verbs “*ealeafcAl ealmuctalla*” /الأفعال المعتلة/: these are the verbs that contain a weak letter. Weak verbs are also classified into three categories (Attia, 2005):

    a. Assimilated “*ealmi~al*” /المثال/: a verb that contains an initial weak letter.
    b. Hollow “*ealajwaf*” /الأجوف/: a verb that contains a middle weak letter.

c. Defective “*eannAqiS*” /الناقص/: a verb that contains a final weak letter.

While the second words category contains three families of words:

- The particular nouns “*ealasmAe ealxASSa*” /الأسماء الخاصة/: these nouns comprise proper nouns, names of cities, names of countries, etc. It also regroups the exclusive nouns “*easmAe ealeisti~nAe*” /أسماء الاستثناء/, the interrogative nouns “*easmAe ealeistifhAm*” /أسماء الاستفهام/, the demonstrative nouns “*easmAe ealei^Ara*” / أسماء الإشارة/, the conditional nouns “*easmAe ea^^art*” /أسماء الشرط/, etc.

- The particles “*ealHurUf*” /الحروف/ likes for example “*HurUfu ealjarri*” /حروف الجر/, “*HurUfu ealjazmi*” / حروف الجزم/, “*HurUfu ealcaTfi*” /حروف العطف/, etc.

- The incomplete verbs “*ealeafcAl eannAqiSa*” /الأفعال الناقصة/: this family of verbs contains the family of verb “*kAda*” /كاد/, the family of verb “*kAna*” /كان/ and the family of verb “*Zanna*” /ظن/.

Finally, after generating a series of AMAUT, their size is about 120 MB. Concerning the number of the entries generated, it’s about 5961 entries, which represent a remarkable reduction of the entries number and makes our approach as one of the best existing approaches in the literature. We could mention that using XMODEL to implement the lexicon could be another advantage that explains the obtained results. To concretize this remarkable implementation, figure 12 shows an AMAUT generated from verbs with the schemes: “*facala*”, “*facila*”, “*facula*” and “*faclala*”.

So related to the AMAUT generated in figure 12, it contains 19 states including 4 accepting states (’19’, ‘16’, ‘17’ and ‘18’) which represent the four schemes. This implementation allows a remarkable reduction concerning the number of the morphological entries and explains the results seen before.

We note that developing the Arabic morphological automata is the proposed approach to develop an Arabic morphological analyzer. In the following paragraph we will present the application of our approach in Arabic morphological analysis.

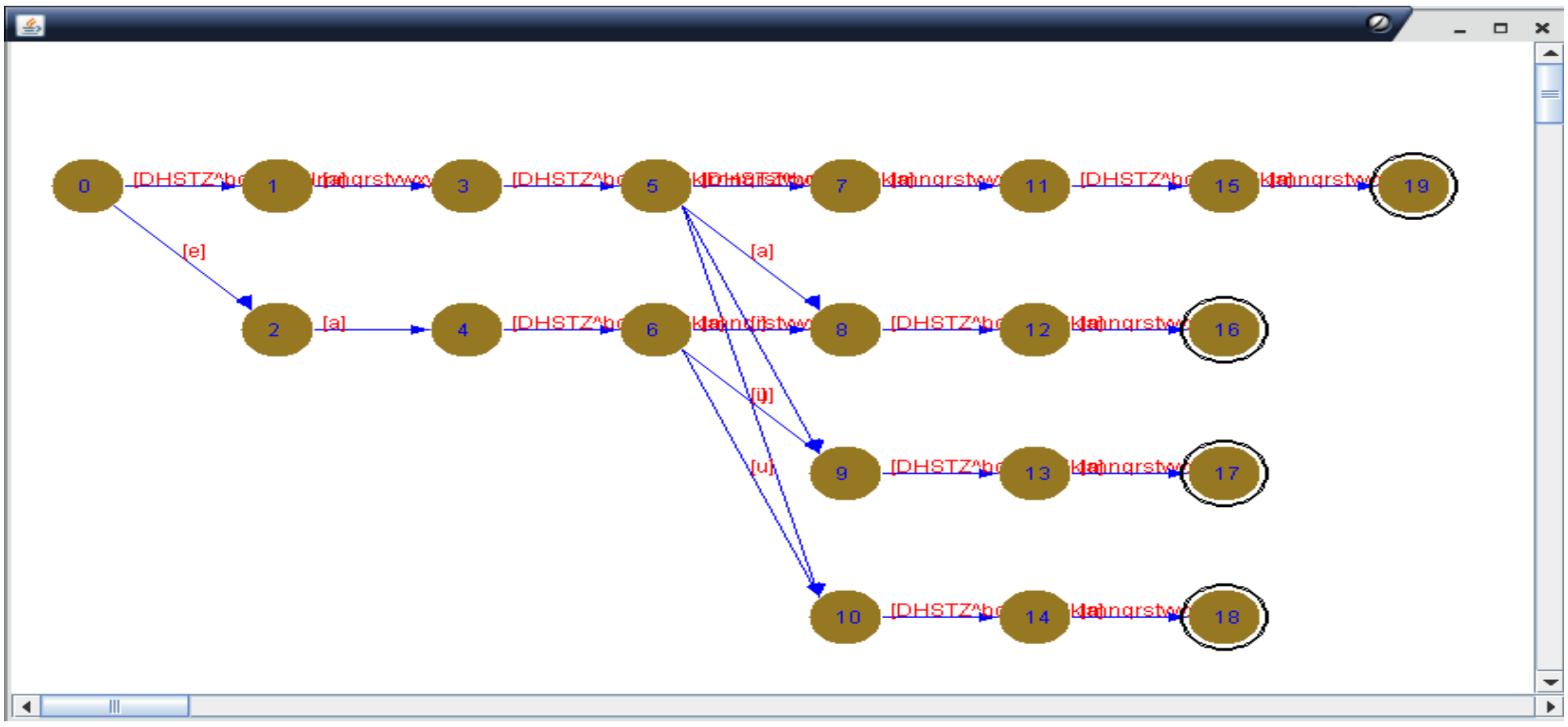

Figure 12. An automaton representing the schemes “*facala*”, “*facila*”, “*facula*” and “*faclala*”

## 6. Application in Arabic Morphological Analysis

In this section, we operate our approach in Arabic morphological analysis. It is based on the Arabic Morphological Automaton (AMAUT) method presented in the previous section. The implementation of our approach has been done using an oriented object framework. It is developed using Java Programming Language and based on a reduced lexicon built using XMODEL language.

The use of AMAUT technology makes our system usable as a generator as well as an analyzer, unlike some morphological analyzers which cannot be converted to generators in a straightforward manner (Sforza, 2000; Buckwalter, 2004; Habash, 2004).

So as to develop an Arabic morphological analyzer and generator, firstly, we used a lexicon built using the XMODEL language integrating the entries suitable for Arabic language. It regroups three packages: morphological components package that contains verbs, nouns, particles and affixes. The second package includes the morphological rules and the last package is concerned with the morphological properties. Secondly, we used a set of Arabic morphological automata each one represents a very specific morphological category. It is considered as the main idea to develop an Arabic morphological analyzer. Finally, we developed a framework handling the lexicon and the morphological automata.

The presented method involves five steps. In this paragraph, we provide a brief description of the principle of this method. As input, the proposed technique accepts an Arabic text. The first step is to apply a tokenization process to the text given. Then, a set of AMAUT are loaded, in a second step. The part-of-speech is determined in the third step. After that, the method determines all possible affixes. Then the next step consists of extracting the morpho-syntactic features according to the valid affixes.

The tokenization process consists of extracting all the words from the text given. A set of Arabic morphological automata are loaded from a package that contains all the implemented Arabic morphological automata. Then, the approach determines which AMAUT is suitable for that word. The result may be one or more AMAUT loaded. Then, the method determines the part-of-speech. If the word analyzed is a noun or a verb, the method determines if it contains a scheme. Then, if it is a verb, the method determines the type of the verb (strong, weak, or incomplete), its tense ("mADI" /ماضي/, "muDAric" /مضارع/ or "eamr" /أمر/), its voice (active or passive), etc. If it is a noun, we determine if it is a derived noun or particular noun. If it is a particle, the method determines if it is a preposition particle /حروف الجر/, conjunction particle /حروف العطف /, etc. After that, the method applied a process of extracting the possible affixes attached to the word analyzed. The next step consists of extracting the morpho-syntactic features according to the valid affixes and the scheme. Additional information is extracted called in our approach morphological descriptors. They describe the word analyzed and they are very useful especially in Natural Language Processing applications. Finally, the morphological analyzer displays the results in a table where each row contains the word analyzed and all the data characterizing this word (see figure 13 and figure 14).

Concerning the morpho-syntactic features given by the morphological analysis using the proposed technique, they are very rich regarding the information given to the user (see figure 13 and 14). It concerns the morphological level, the syntactic and semantic level which makes the richness of our system compared to the others system (see the Evaluation section). Here are some important features which will be given by the system.

- The word gender: masculine or feminine.
- The word person: first, second or third person.
- The word number: singular, dual or plural.
- The word case: "*marfUc*" (مرفوع), "*manSUb*" (منصوب), "*majrUr*" (مجرور), "*majzUm*" (مجزوم).
- The type of the word: verb, noun or particle.
- If the word is a verb, we give its tense: present ("*ealmuDAric*": المضارع), past ("*ealmADI*": الماضي) or imperative ("*ealeamr*": الأمر). We also give its voice: active or passive.
- The scheme of the word is given if available.

To concretize these obtained results, we analyze some examples of Arabic verbs and nouns using the proposed technique. These examples are taken from a standard input text provided by ALECSO (Arab League, Educational, Cultural and Scientific Organization) which organized a competition in April 2009 of the Arabic Analyzers in Damascus. The standard input text provided by ALECSO is unvocalized, in this test, we used a vocalized version. This standard input text is provided in this file: http://www.alecso.org.tn/images/stories/OULOUM/MOHALLILAT%20SARFIADAMAS2009/020%20 NIZAR.html.

Lexical Table of File : arab files\dfa01.txt

| A | B | C | D | E | F | G | H | I | J |
|---|---|---|---|---|---|---|---|---|---|
| Morphological | Original Scheme | Scheme | Gender | Person | Number | Properties | Morphological Descriptors | Prefixes | Suffixes |
| yatadaHrajAni | [tafaclala], | [] | GMa | ,Pr3 | ,NDl | Strong Verb,MOD,ACT, | Raf, | [y] | [Ani] |
| eaSTaffa | [eifcalla], | [] | GFe,GMa | ,Pr1 | ,NSg | Strong Verb,ACT,MOD, | Def,NaS, | [e] | [a] |
| eaSTaffa | [eifcalla], | [] | GFe,GMa | ,Pr1 | ,NSg | Strong Verb,ACT,MOD, | Def,NaS, | [e] | [a] |
| eaSTaffu | [eifcalla], | [] | GFe,GMa | ,Pr1 | ,NSg | Strong Verb,ACT,MOD, | Def,Raf, | [e] | [u] |
| yacuddu | [cadda], | [] | GMa | ,Pr3 | ,NSg | Incomplete Verb,MOD, | Def,Raf, | [y] | [u] |
| yacidu | [wacala, wacila, wacula], | [] | GMa | ,Pr3 | ,NSg | Weak Verb,ACT,MOD, | Def,Raf, | [y] | [u] |
| yucda | [facA, faciya], | [] | GMa | ,Pr3 | ,NSg | Weak Verb,PAS,MOD, | NaS,Jaz, | [y] | [a] |
| yari~a | [wacala, wacila, wacula], | [] | GMa | ,Pr3 | ,NSg | Weak Verb,ACT,MOD, | Def,NaS, | [y] | [a] |
| yar~i | [eafcA], | [] | GMa | ,Pr3 | ,NSg | Weak Verb,ACT,MOD, | NaS,Jaz, | [y] | [i] |
| yur~a | [facA, faciya], | [] | GMa | ,Pr3 | ,NSg | Weak Verb,PAS,MOD, | NaS,Jaz, | [y] | [a] |
| yari~u | [wacala, wacila, wacula], | [] | GMa | ,Pr3 | ,NSg | Weak Verb,ACT,MOD, | Def,Raf, | [y] | [u] |
| eastaqbila | [eistafcala], | [] | GFe,GMa | ,Pr1 | ,NSg | Strong Verb,ACT,MOD, | Def,NaS, | [e] | [a] |
| eastaqbilu | [eistafcala], | [] | GFe,GMa | ,Pr1 | ,NSg | Strong Verb,ACT,MOD, | Def,Raf, | [e] | [u] |
| tuSAfiHa | [fAcala], | [] | GMa,GFe | ,Pr2,Pr3 | ,NSg | Strong Verb,ACT,MOD, | Def,NaS, | [t] | [a] |
| tuSAfiHa | [fAcala], | [] | GMa,GFe | ,Pr2,Pr3 | ,NSg | Strong Verb,ACT,MOD, | Def,NaS, | [t] | [a] |
| tuSAfiHu | [fAcala], | [] | GMa,GFe | ,Pr2,Pr3 | ,NSg | Strong Verb,ACT,MOD, | Def,Raf, | [t] | [u] |
| eabtacidu | [eiftacala], | [] | GFe,GMa | ,Pr1 | ,NSg | Strong Verb,ACT,MOD, | Def,Raf, | [e] | [u] |
| eabtacida | [eiftacala], | [] | GFe,GMa | ,Pr1 | ,NSg | Strong Verb,ACT,MOD, | Def,NaS, | [e] | [a] |
| yankami^a | [einfacala], | [] | GMa | ,Pr3 | ,NSg | Strong Verb,ACT,MOD, | Def,NaS, | [y] | [a] |
| yankami^u | [einfacala], | [] | GMa | ,Pr3 | ,NSg | Strong Verb,ACT,MOD, | Def,Raf, | [y] | [u] |
| tusarbila | [faclala], | [] | GMa,GFe | ,Pr2,Pr3 | ,NSg | Strong Verb,ACT,MOD, | Def,NaS, | [t] | [a] |
| tusarbila | [faclala], | [] | GMa,GFe | ,Pr2,Pr3 | ,NSg | Strong Verb,ACT,MOD, | Def,NaS, | [t] | [a] |
| tusarbilu | [faclala], | [] | GMa,GFe | ,Pr2,Pr3 | ,NSg | Strong Verb,ACT,MOD, | Def,Raf, | [t] | [u] |
| tusarbilu | [faclala], | [] | GMa,GFe | ,Pr2,Pr3 | ,NSg | Strong Verb,ACT,MOD, | Def,Raf, | [t] | [u] |
| lAHaZat | [fAcala], | [] | GFe | , Pr3 | ,NSg | Strong Verb,ACT,MAD, | | [] | [at] |
| lAHaZtu | [fAcala], | [] | GFe,GMa | ,Pr1 | ,NSg | Strong Verb,MAD,ACT, | | [] | [tu] |
| lAHaZta | [fAcala], | [] | GMa | ,Pr2 | ,NSg | Strong Verb,MAD,ACT, | | [] | [ta] |

Figure 13. A morphological analysis of some verbs using the proposed technique

Figure 13 and 14 show the morphological analysis results of some words analyzed using the proposed technique. As discussed before, the analyzer displays the Part-of-speech (verb, noun or particle), the original scheme is displayed in column B because Arabic has this particularity which is summarized in that some words might be conjugated forms of other words like "*afcalu*", "*afcilu* ", "*afculu*", these three words are all conjugated forms of "*facala*". The gender (masculine or feminine) is displayed in column D, the person (first, second or third person) is displayed in column E, the number (singular, dual or plural) is displayed in column F. For the column G, it concerns some properties that characterize the word analyzed and they are very useful to the user. Some morphological descriptors are displayed in column H. Finally, the column I and J show the affixes attached to the word.

Finally, the proposed technique for Arabic morphological analysis has many advantages such as:

- The separation between the linguist and the developer task.
- We can also reuse our programs in future works.
- Development standardization means in our application that we have build all the applications with the same standards.
- The facility of maintenance: it's easy to add some new features or morphological characteristics to the presented system if the user or the linguist needs them for his Arabic morphological analysis. It's also easy to extend our system to include some new works related to Arabic NLP such as information retrieval, syntactic and semantic analyzers, correction and generation of Arabic texts.

Lexical Table of File : arab files\ConstructionNomD.txt

| A | B | C | D | E | F | G | H | I | J |
|---|---|---|---|---|---|---|---|---|---|
| Morpho Cmp | Original Scheme | Scheme | Gender | Person | Number | Properties | Morphological Descriptors | Prefixes | Suffixes |
| xArijUna | [facala, facila, facula], | [fAcil] | GMa | | ,NPl | Derived Noun,NomDAFP,NomDAD,accepteSC,acceptel,NomDAT,NomDAMP, | efc,Raf, | [] | [Una] |
| Zabyatu | Not exist, | [facl, faclal] | GFe | | | Derived Noun,NomDAFP,NomDAT, | mmr,maS,Def,Raf, | [] | [u, at, at] |
| Zabyatun | Not exist, | [facl, faclal] | GFe | | | Derived Noun,NomDAFP,NomDAT, | mmr,maS,Ind,Raf, | [] | [un, at, at] |
| Zabyatin | Not exist, | [facl, faclal] | GFe | | | Derived Noun,NomDAFP,NomDAT, | mmr,maS,Ind,KaS, | [] | [in, at, at] |
| mucTayAtin | [eafcala], | [mufcal] | GFe | | ,NPl | Derived Noun,NomDAFP,NomDAD,accepteSC,acceptel,NomDAT,NomDAMP, | emf,mmi1,Ind,KaS, | [] | [in, At] |
| mucTiyAti | [eafcala], | [mufcil] | GFe | | ,NPl | Derived Noun,NomDAFP,NomDAD,accepteSC,acceptel,NomDAT,NomDAMP, | efc,Def,KaS, | [] | [i, At] |
| naZrati | Not exist, | [facl, faclal] | GFe | | | Derived Noun,NomDAFP,NomDAT, | mmr,maS,Def,KaS, | [] | [i, at, at] |
| naZratan | Not exist, | [facl, faclal] | GFe | | | Derived Noun,NomDAFP,NomDAT, | mmr,maS,Ind,NaS, | [] | [an, at, at] |
| naZratun | Not exist, | [facl, faclal] | GFe | | | Derived Noun,NomDAFP,NomDAT, | mmr,maS,Ind,Raf, | [] | [un, at, at] |
| maktabi | Not exist, | [faclal, mafcal] | GFe,GMa | | | Derived Noun,acceptel,NomDAT,NomDAFP,accepteSC,NomDAD, | maS,mmi,Def,KaS, | [] | [i] |
| maktabun | Not exist, | [faclal, mafcal] | GFe,GMa | | | Derived Noun,acceptel,NomDAT,NomDAFP,accepteSC,NomDAD, | maS,mmi,Ind,Raf, | [] | [un] |
| maktabin | Not exist, | [faclal, mafcal] | GFe,GMa | | | Derived Noun,acceptel,NomDAT,NomDAFP,accepteSC,NomDAD, | maS,mmi,Ind,KaS, | [] | [in] |
| mlnAean | Not exist, | [flIAl] | GMa | | | Derived Noun,acceptel,NomDAFP,accepteSC,NomDAD, | maS,Ind,NaS, | [] | [an] |
| mlnAea | Not exist, | [flIAl] | GMa | | | Derived Noun,acceptel,NomDAFP,accepteSC,NomDAD, | maS,Def,NaS, | [] | [a] |
| mlnAeu | Not exist, | [flIAl] | GMa | | | Derived Noun,acceptel,NomDAFP,accepteSC,NomDAD, | maS,Def,Raf, | [] | [u] |
| turA~a | Not exist, | [fucAl] | GMa | | | Derived Noun,acceptel,NomDAFP,accepteSC,NomDAD, | maS,Def,NaS, | [] | [a] |
| turA~in | Not exist, | [fucAl] | GMa | | | Derived Noun,acceptel,NomDAFP,accepteSC,NomDAD, | maS,Ind,KaS, | [] | [in] |
| turA~un | Not exist, | [fucAl] | GMa | | | Derived Noun,acceptel,NomDAFP,accepteSC,NomDAD, | maS,Ind,Raf, | [] | [un] |
| biTAqAti | Not exist, | [ficAl, ficAl] | GMa,GFe | | ,NSg,NPl | Derived Noun,NomDAFP,NomDAD,accepteSC,acceptel,NomDAT, | maS,jtS,Def,KaS, | [] | [i, At] |
| biTAqAti | Not exist, | [ficAl, ficAl] | GMa,GFe | | ,NSg,NPl | Derived Noun,NomDAFP,NomDAD,accepteSC,acceptel,NomDAT, | maS,jtS,Def,KaS, | [] | [i, At] |
| biTAqAti | Not exist, | [ficAl, ficAl] | GMa,GFe | | ,NSg,NPl | Derived Noun,NomDAFP,NomDAD,accepteSC,acceptel,NomDAT, | maS,jtS,Def,KaS, | [] | [i, At] |
| yusrin | Not exist, | [fucl, fucl] | GMa,GFe | | | Derived Noun,NomDAT,acceptel,NomDAFP,accepteSC,NomDAD, | Smb,maS,Ind,KaS, | [] | [in] |
| yusri | Not exist, | [fucl, fucl] | GMa,GFe | | | Derived Noun,NomDAT,acceptel,NomDAFP,accepteSC,NomDAD, | Smb,maS,Def,KaS, | [] | [i] |
| yusrun | Not exist, | [fucl, fucl] | GMa,GFe | | | Derived Noun,NomDAT,acceptel,NomDAFP,accepteSC,NomDAD, | Smb,maS,Ind,Raf, | [] | [un] |
| naqba | Not exist, | [facl] | GFe | | | Derived Noun,NomDAFP,NomDAT, | mmr,Def,NaS, | [] | [a] |
| naqbin | Not exist, | [facl] | GFe | | | Derived Noun,NomDAFP,NomDAT, | mmr,Ind,KaS, | [] | [in] |
| naqbu | Not exist, | [facl] | GFe | | | Derived Noun,NomDAFP,NomDAT, | mmr,Def,Raf, | [] | [u] |
| wahmun | Not exist, | [facl] | GFe | | | Derived Noun,NomDAFP,NomDAT, | mmr,Ind,Raf, | [] | [un] |
| wahmu | Not exist, | [facl] | GFe | | | Derived Noun,NomDAFP,NomDAT, | mmr,Def,Raf, | [] | [u] |
| wahmin | Not exist, | [facl] | GFe | | | Derived Noun,NomDAFP,NomDAT, | mmr,Ind,KaS, | [] | [in] |
| manna | Not exist, | [facl] | GFe | | | Derived Noun,NomDAFP,NomDAT, | mmr,Def,NaS, | [] | [a] |
| manni | Not exist, | [facl] | GFe | | | Derived Noun,NomDAFP,NomDAT, | mmr,Def,KaS, | [] | [i] |
| wirA~ata | Not exist, | [ficAl] | GFe | | | Derived Noun,NomDAFP,NomDAT, | maS,Def,NaS, | [] | [a, at, at] |
| wirA~ati | Not exist, | [ficAl] | GFe | | | Derived Noun,NomDAFP,NomDAT, | maS,Def,KaS, | [] | [i, at, at] |
| wirA~atin | Not exist, | [ficAl] | GFe | | | Derived Noun,NomDAFP,NomDAT, | maS,Ind,KaS, | [] | [in, at, at] |
| kitAbin | Not exist, | [ficAl, ficAl] | GFe,GMa | | ,NSg,NPl | Derived Noun,NomDAT,acceptel,NomDAFP,accepteSC,NomDAD, | maS,jtS,Ind,KaS, | [] | [in] |
| kitAba | Not exist, | [ficAl, ficAl] | GFe,GMa | | ,NSg,NPl | Derived Noun,NomDAT,acceptel,NomDAFP,accepteSC,NomDAD, | maS,jtS,Def,NaS, | [] | [a] |
| kuttAbin | Not exist, | [fuccAl] | GMa | | ,NPl,NSg | Derived Noun,accepteSC,acceptel, | Sif,efc,Smb,Ind,KaS, | [] | [in] |

Figure 14. Morphological analysis of some nouns using the proposed technique

## 7. Evaluation

Firstly, we mention that a standard annotated corpus for Arabic language is not yet available, for this reason the evaluation process will be difficult. To evaluate the presented approach for morphological analysis, we use a corpus taken from a standard input text provided by ALECSO which organized a competition in April 2009 of the Arabic Morphological Analyzers in Damascus. In this section, we evaluate Xerox Arabic Morphological Analyzer, the Arabic Morphological Analyzer by Otakar Smrz and our approach. We have chosen Xerox Morphological Analyzer and the Morphological Analyzer by Otakar Smrz because they are one of the best known morphological analyzers for MSA and they are available and well documented.

On the one hand, the first remark when we compare the three morphological analyzers is about the richness of information giving by each one. Using morphological automaton approach and XMODEL language to represent morphological knowledge, our morphological analyzer gives more information about each word analyzed and more precision compared to Xerox Arabic Morphological Analyzer and Smrz's Analyzer. To clarify this point, we select some Arabic words from ALECSO input text and try to analyze them using the three morphological analyzers.

Table 4, 5 and 6 show the results of ten different Arabic words analyzed using the three morphological analyzers. We note that the proposed approach provides more information and more precision about the word analyzed compared to the others. Thanks to the power of the morphological automaton for Arabic and XMODEL language which permit to represent the morphological knowledge in an optimal way. This advantage will be very useful especially in the future works which will be done later. It should be noted that the presented work could provide more information about the word analyzed according to the user needs.

Table 4. Words analyzed using Xerox Arabic Morphological Analyzer

| The word | Morphological Analysis using Xerox Arabic Morphological Analyzer |
|---|---|
| صِفْرٌ [Sifrun] | CiCoC Noun +N Indef Nom |
| خَارِجُونَ [xArijUna] | CACiC participle Active +U3na Masc Plur Nom |
| مُرْتَدِّي [murtaddI] | muCtaCaC Participle Passive +I3 Ma Plur Acc/Gen Possessive |
| فُصِلْتُ [fuSiltu] | +tu 1stPer Masc/Fem Sing CuCiC Verb |
| أُخْرِجْتُمَا [euxrijtumA] | uCoCiC Verb +tumA 2ndPer Masc/Fem Dual |
| مَعَ [maca] | maEa Funcwa |
| أَمَامَ [eamAma] | CaCAC Noun +a Def Acc |
| العَاشِرَ [ealcA^ira] | al Article CACiC Noun +a Def Acc |
| بِهِمَا [bihimA] | bi +himA Funcwa |
| يُجَادِلُونَ [yujAdilUna] | yu Imperfect Prefix CACiC Verb +Una Indicative 3rdPer Masc Plur |

Table 5. Words analyzed using ElixirFM

| The word | Morphological Analysis using ElixirFM |
|---|---|
| صِفْرٌ [Sifrun] | FiCL-un noun, singular, nominative, indefinite |
| خَارِجُونَ [xArijUna] | FāCiL-ūna adjective, masculine, plural, nominative, indefinite |
| مُرْتَدِّي [murtaddI] | muFtaCL-ī noun, plural, genitive, reduced/construct |
| فُصِلْتُ [fuSiltu] | FuCiL-tu perfective verb, passive, first person, singular |
| أُخْرِجْتُمَا [euxrijtumA] | uFCiL-tumā perfective verb, passive, second person, dual |
| مَعَ [maca] | inflected preposition, accusative |
| أَمَامَ [eamAma] | FaCāL-a inflected preposition, accusative |
| العَاشِرَ [ealcA^ira] | al-FāCiL-a adjective, masculine, singular, accusative, definite |
| بِهِمَا [bihimA] | This word is divided into « bi » and « himA » |
| يُجَادِلُونَ [yujAdilUna] | yu-FāCiL-ūna imperfective verb, indicative, active, third person, masculine, plural |

Table 6. Words analyzed using our morphological analysis approach

| The word | Morphological Analysis using our Approach |
|---|---|
| صِفْرٌ [Sifrun] | Gma Particular Noun V0 Ind Raf [un] |
| خَارِجُونَ [xArijUna] | facala facila facula fAcil Gma NPl Derived Noun accepteSC accepteI efc Raf [Una] |
| مُرْتَدِّي [murtaddI] | eifcalla mufcallI Gma Pr1 NDl NSg Derived Noun accepteSC accepteI emf mmi8 KaS [I] |
| فُصِلْتُ [fuSiltu] | facula facala facila Gfe Gma Pr1 NSg Strong Verb MAD PAS [tu] |
| أُخْرِجْتُمَا [euxrijtumA] | eafcala Gfe Gma Pr2 NDl Strong Verb MAD PAS [tumA] |
| مَعَ [maca] | Particle accepteI zam mak Def NaS [a] |
| أَمَامَ [eamAma] | Particle accepteI mak Def NaS [a] |
| العَاشِرَ [ealcA^ira] | Particular Noun V10 Def NaS [eal] [a] |
| بِهِمَا [bihimA] | Gfe Gma Pr3 NPl KaS [bi] [himA] |
| يُجَادِلُونَ [yujAdilUna] | fAcala Gma Pr3 NPl Strong Verb MOD ACT Raf [y] [Una] |

On the other hand, let us see the evaluation process from another view. We have selected corpora of 975 words taken from ALECSO input text containing different part-of-speech (verbs, nouns and particles). Then, we tested them on each morphological analyzer, after that we draw a detailed analysis for the three analyzers. The corpora contain 975 words divided into 481 nouns, 362 verbs and 132 particles. Table 7 shows number of words which are not found when they are analyzed using the three morphological analyzers:

Table 7. The evaluation process results

| Part-of-Speech | The number | Xerox Morphological Analyzer | ElixirFM | Our System |
|---|---|---|---|---|
| Nouns | 481 | 39 | 76 | 21 |
| Verbs | 362 | 16 | 21 | - |
| Particles | 132 | 29 | 55 | - |
| Total | 975 | 84 | 152 | 21 |

The proposed technique for morphological analysis can reach an average of performance around 90% which will make it one of the best existing approaches for Arabic morphological analysis and it will be very useful for the next future works to be done in NLP. We note that an update of the presented lexicon could resolve these errors seen in table 7.

## 8. Discussion

To compare the proposed technique for Arabic morphological analysis to the other existing approaches, the task becomes difficult because, on the one hand, there is no standard to make the comparison. On the other hand, each system has its own target. For this reason, each approach has some advantages and disadvantages comparing to the others.

The proposed technique has some advantages comparing to the other approaches. These advantages are:

- Our approach can be used in both analysis and generation
- The use of morphological automaton approach makes the system efficient and very fast
- Our new and innovative language (XMODEL) used to represent the morphological knowledge and the use of morphological automaton permit to avoid a huge problems of ambiguity in Arabic which the most approaches can't resolve
- The use of XMODEL language permit to reduce the number of the entries in the lexicon which present a big problem in the other approaches
- The updating process of the Arabic morphological database is very easy to develop. This advantage makes our system very flexible
- The major advantage of our system is that it allows, on the one hand, giving the affixes, the stem and the scheme for any word given. On the other hand, it gives morpho-syntactic features about the word analyzed and an additional morphological descriptors which permit to characterize every Arabic word

This work has some disadvantages compared to some other systems. Firstly, it can't handle undiacritized texts. Secondly, it doesn't provide an English glossary and finally, it handles words which do not exist. To clarify the last disadvantage, let us take an example of some invalid words analyzed with our system.

Figure 15 shows five invalid words analyzed as they are valid words in Arabic. So the next works to be done is to solve this problem. To solve it, we will keep the current design of morphological analysis. Our idea is to use a lexicon of any existing morphological analyzer (for example: Buckwalter's analyzer, Xerox analyzer etc.) and eliminate the invalid words generated by the morphological automaton. This operation will reduce the number of the invalid words. But, this kind of problems isn't very serious to take in consideration, because Arabic Language is very rich in words and every year there is some news words added to this language.

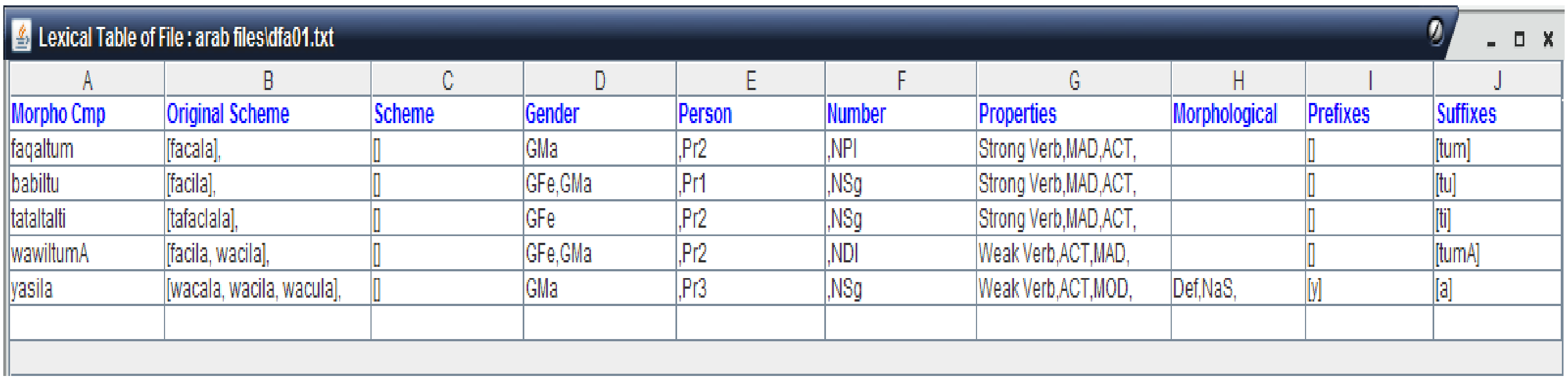

Lexical Table of File : arab files\dfa01.txt

| A | B | C | D | E | F | G | H | I | J |
|---|---|---|---|---|---|---|---|---|---|
| Morpho Cmp | Original Scheme | Scheme | Gender | Person | Number | Properties | Morphological | Prefixes | Suffixes |
| faqaltum | [facala], | [] | GMa | ,Pr2 | ,NPl | Strong Verb,MAD,ACT, | | [] | [tum] |
| babiltu | [facila], | [] | GFe,GMa | ,Pr1 | ,NSg | Strong Verb,MAD,ACT, | | [] | [tu] |
| tataltalti | [tafaclala], | [] | GFe | ,Pr2 | ,NSg | Strong Verb,MAD,ACT, | | [] | [ti] |
| wawiltumA | [facila, wacila], | [] | GFe,GMa | ,Pr2 | ,NDl | Weak Verb,ACT,MAD, | | [] | [tumA] |
| yasila | [wacala, wacila, wacula], | [] | GMa | ,Pr3 | ,NSg | Weak Verb,ACT,MOD, | Def,NaS, | [y] | [a] |

Figure 15. Some invalid words analyzed by our approach

## 9. Conclusion

We have described an approach for Arabic morphological analysis. It is called the Arabic Morphological Automaton (AMAUT). We have evaluated the presented approach using Xerox Arabic Morphological Analyzer and Arabic Morphological Analyzer by Otakar Smrz because they are considered as the most referenced approaches for Arabic morphological analysis and they are available for research and evaluation. The use of the Arabic morphological automaton makes the morphological analyzer efficient and very fast. Concerning the development of the lexicon, we have used XMODEL language for representing, designing and implementing the lexical resource. Our approach has another advantage because it’s developed using Java language and XML technology which makes the system portable and reusable.

## Appendix 1. Features signification in morphological analysis

| **Feature** | **Description** |
|---|---|
| Gfe | Feminine |
| Gma | Masculine |
| Def | Defined |
| Ind | Undefined |
| NaS | manSUb « منصوب » |
| KaS | majrUr « مجرور » |
| Raf | marfUc « مرفوع » |
| Jaz | majzUm « مجزوم » |
| NSg | Singular |
| NDl | Dual |
| NPl | Plural |
| Pr1 | First Person |
| Pr2 | Second Person |
| Pr3 | Third Person |
| MOD | ealmuDAric « المضارع » |
| MAD | ealmADI « الماضي » |
| ACT | Active |
| PAS | Passive |
| AccepteSC | Accept Case Suffixes |
| Efc | Eismu fAcil « اسم فاعل » |
| Emf | Eismu mafcUl « اسم مفعول » |
| Mmi | maSdar mImI « اسم ميم » |
| Zam | Zarfu zamAn « ظرف زمان » |
| Mak | Zarfu makAn « ظرف مكان » |
| Mmr | maSdar_ealmarrat « مصدر المرة » |
| MaS | maSdar « مصدر » |
| JtS | jamcu taksIr li Sifatin « جمع تكسير لصفة » |
| Smb | SIgatu ealmubalagati « صيغة المبالغة » |
| Sif | Sifatun « صفة » |

# Appendix 2. Letter mappings

| | | | | | | | | | | |
|---|---|---|---|---|---|---|---|---|---|---|
| ا | : | A | | س | : | s | | ك | : | k |
| ب | : | B | | ش | : | ^ | | ل | : | l |
| ت | : | T | | ص | : | S | | م | : | m |
| ث | : | ~ | | ض | : | D | | ن | : | n |
| ج | : | J | | ط | : | T | | هـ | : | h |
| ح | : | H | | ظ | : | Z | | و | : | w |
| خ | : | X | | ع | : | c | | ي | : | y |
| د | : | D | | غ | : | g | | ى | : | A |
| ذ | : | V | | ف | : | f | | ة | : | t |
| ر | : | r | | ق | : | q | | ء | : | e |
| ز | : | z | | | | | | | | |

# Appendix 3. The English translation of Arabic words

| **The Arabic Word** | **Transliteration** | **English Translation** |
|---|---|---|
| الأفعال الصحيحة | ealeafcAl eaSSaHiyHa | Strong verbs |
| الأفعال المعتلة | ealeafcAl ealmuctalla | Weak verbs |
| الأفعال الناقصة | ealeafcAl eannAqiSa | Defective verbs |
| المثال | ealmi~al | Assimilated |
| الأجوف | Ealajwaf | Hollow |
| الناقص | EannAqiS | Defective |
| الأسماء الخاصة | ealasmAe ealxASSa | Particular nouns |
| الأسماء المشتقة | ealaSmAe ealmu^taqqa | Derived nouns |
| أسماء الاستفهام | easmAe ealeistifhAm | Interrogation nouns |
| أسماء الإشارة | easmAe ealei^Ara | Demonstrative nouns |
| أسماء الشرط | easmAe ea^^art | Condition nouns |
| حروف الجر | HurUfu ealjarri | Preposition particles |
| حروف العطف | HurUfu ealcaTfi | Conjunction particles |
| كان | KAna | Was |
| ظن | Zanna | To think |
| مرفوع | MarfUc | Nominative case |
| منصوب | ManSUb | Accusative case |
| مجرور | MajrUr | Genitive case |
| مجزوم | MajzUm | Jussive case |
| المضارع | ealmuDAric | The Imperfect |
| الماضي | EalmADI | The Perfect |
| الأمر | Ealeamr | The Imperative |
| صِفْرٌ | Sifrun | Zero |
| مَعَ | Maca | With |
| أَمَامَ | EamAma | In front |
| العَاشِرَ | ealcA^ira | The tenth |